\documentclass[journal]{IEEEtran}
\usepackage{amssymb}
\usepackage{amsfonts}

\usepackage{booktabs} 
\usepackage{multirow}
\usepackage{makecell}

\usepackage{algorithm}
\usepackage{algpseudocode}  % 推荐比 algorithmic 更强大
\usepackage{caption}

\usepackage{cite}
\ifCLASSINFOpdf
  \usepackage[pdftex]{graphicx}
\else
\fi
\usepackage[cmex10]{amsmath}
\usepackage{array}
\usepackage{fixltx2e}
\usepackage{url}
\begin{document}

% \title{Synergistic Prompting and Selective Consolidation for Incremental Face Presentation Attack Detection}
% \title{SynapseVLM: Synergistic Adaptation of Vision-Language Models for Lifelong Face PAD}
\title{Steering Vision-Language Pre-trained Models for Incremental Face Presentation Attack Detection}

\author{Haoze~Li,~\IEEEmembership{}
        Jie~Zhang,~\IEEEmembership{Member,~IEEE,}
        Guoying~Zhao,~\IEEEmembership{Fellow,~IEEE,}
        Stephen~Lin,~\IEEEmembership{Member,~IEEE,}
        and~Shiguang~Shan,~\IEEEmembership{Fellow,~IEEE}% <-this % stops a space
\thanks{Haoze Li is with the School of Computer Science, China University of Geosciences, Wuhan 430074, China (e-mail: haozeli123@163.com).}
        
\thanks{Jie Zhang, Shiguang Shan are with the State Key Laboratory of AI Safety, Institute of Computing Technology, Chinese Academy of Sciences (CAS), Beijing 100190, China, and also with the University of China Academy of Sciences, Beijing 100049, China (e-mail: zhangjie@ict.ac.cn;
sgshan@ict.ac.cn).}% <-this % stops a space
\thanks{Guoying Zhao is with the Center for Machine Vision and Signal Analysis, University of Oulu, 90014 Oulu, Finland (e-mail: guoying.zhao@oulu.fi).}% <-this % stops a space
\thanks{Stephen Lin is with Microsoft Research Asia, Beijing 100080, China (e-mail: stevelin@microsoft.com).}
}


\markboth{Journal of \LaTeX\ Class Files,~Vol.~13, No.~9, September~2014}%
{Shell \MakeLowercase{\textit{et al.}}: Bare Demo of IEEEtran.cls for Journals}



\maketitle


\begin{abstract}
Face Presentation Attack Detection (PAD) demands incremental learning (IL) to combat evolving spoofing tactics and domains. Privacy regulations, however, forbid retaining past data, necessitating rehearsal-free IL (RF-IL). Vision-Language Pre-trained (VLP) models, with their prompt-tunable cross-modal representations, enable efficient adaptation to new spoofing styles and domains. Capitalizing on this strength, we propose \textbf{SVLP-IL}, a VLP-based RF-IL framework that balances stability and plasticity via \textit{Multi-Aspect Prompting} (MAP) and \textit{Selective Elastic Weight Consolidation} (SEWC). MAP isolates domain dependencies, enhances distribution-shift sensitivity, and mitigates forgetting by jointly exploiting universal and domain-specific cues. SEWC selectively preserves  critical weights from previous tasks, retaining essential knowledge while allowing flexibility for new adaptations. Comprehensive experiments across multiple PAD benchmarks show that SVLP-IL significantly reduces catastrophic forgetting and enhances performance on unseen domains. SVLP-IL offers a privacy-compliant, practical solution for robust lifelong PAD deployment in RF-IL settings.
\end{abstract}


% \fi
\IEEEpeerreviewmaketitle

\section{Introduction}
\IEEEPARstart{F}{ace recognition (FR)} systems have become deeply integrated into modern society, offering convenient and efficient identity verification for critical applications ranging from smartphone security to financial transactions and public safety. However, their widespread application also makes them a prime target for malicious attacks. Among these, Presentation Attacks (PAs), where adversaries employ physical artifacts like printed photos, video replays, or even 3D masks to deceive the system~\cite{Yu2023,Kong2022Dig,Ramach2017}, constitute one of the most direct and pervasive threats. To ensure the trustworthiness and security of FR systems, the development of robust Presentation Attack Detection (PAD) mechanisms has become an urgent and indispensable task for both academia and industry.

Deploying robust Presentation Attack Detection (PAD) systems in dynamic, open-world environments presents a fundamental and persistent challenge. Attack types continually evolve with technological advancements, progressing from simple printouts to high-resolution video replays and sophisticated silicone masks. Simultaneously, operational environments exhibit constant variation across lighting conditions, camera sensors, and user demographics~\cite{li2018unsupervised}. This inherent dynamism necessitates PAD systems capable of lifelong learning, which continuously adapt throughout their operational lifecycle. Traditional one-time training paradigms prove empirically fragile in this context. While retraining models from scratch is computationally prohibitive, merely fine-tuning on new data induces catastrophic forgetting~\cite{kirkpatrick2017}, eroding the model's ability to detect prior attack types. Consequently, adopting an Incremental Learning (IL) paradigm becomes essential.

In practice, however, IL for PAD confronts a near-insurmountable obstacle: data privacy. Stringent regulations like GDPR  \cite{gdpr2018general}, coupled with the inherent sensitivity of facial biometrics, typically prohibit storing or reusing historical user data for model updates unless strict compliance measures and explicit consent are ensured. This constraint necessitates solutions under the demanding paradigm of Rehearsal-Free Incremental Learning (RF-IL). In the RF-IL setting, where the model cannot access past tasks, the risk of catastrophic forgetting is profoundly exacerbated. The model must strike a delicate balance within the stability-plasticity dilemma: preserving stable knowledge of previous tasks while remaining plastic enough to learn new ones. Although recent RF-IL methods have explored solutions at the visual feature level, such as parameter isolation or feature regularization~\cite{cai2023, wang2024multi}, they usually degenerate when faced with significant domain shifts. When new spoofing types are visually distinct from old ones, relying exclusively  on visual cues proves inadequate for building robust lifelong PAD systems.

To overcome the limitations of purely visual approaches, we posit that incorporating richer cross-modal prior knowledge is essential. Vision-Language Pre-trained (VLP) models, such as CLIP\cite{radford2021}, which are pre-trained on vast corpora of image-text pairs, offer a transformative solution. By mapping images and text to a shared semantic space, VLP models encode not only object identity,  representing the ``what'', but also appearance attributes,  capturing the ``how''. This capability  proves critical for PAD. Natural language can precisely articulate spoof artifact characteristics, such as ``moiré patterns from screens,'' ``color distortions in printouts,'' or ``rigid mask edges.'' By leveraging such semantic guidance, VLP models have the potential to learn more generalizable and transferable spoofing cues, enabling more feasible adaptation to novel attack types and domains.

Yet, it is non-trivial to directly integrate VLP models into an incremental learning framework. A core difficulty lies in managing the adaptability of the large, powerful VLP backbone. Fully freezing the backbone to ensure stability would prevent the model from learning new domain-specific visual features, leading to poor adaptation. Conversely, allowing unrestricted fine-tuning for maximum plasticity would rapidly corrupt the valuable general knowledge from VLP pre-training and the specific knowledge from previous PAD tasks, resulting in even more severe forgetting. 
%% NEW: begin
Equally problematic, applying uniform all-parameter regularization across a long task sequence, such as dense EWC-style penalties~\cite{kirkpatrick2017}, accumulates constraints over time. In our rehearsal-free PAD setting with a large VLP backbone, penalizing all network parameters across a long sequence over-regularizes and hinders adaptation.
%% NEW: end

To effectively navigate this stability-plasticity trade-off, we propose Steering Vision-Language Pre-trained models for Incremental Learning (SVLP-IL), a rehearsal-free IL framework designed specifically for PAD. SVLP-IL steers the VLP model's adaptation process through two synergistic components,  Multi-Aspect Prompting (MAP) and Selective Elastic Weight Consolidation (SEWC). MAP is a parameter-efficient technique operating at the VLP model's input level. Unlike previous methods\cite{wang2022_sprompts} that learn isolated prompts per domain, MAP employs a hierarchical prompting system to explicitly decompose knowledge. This system captures both universal, cross-domain spoofing cues, such as the two-dimensional flatness common to artifacts, and domain-specific characteristics, including unique moiré patterns from specific screen types. This dual-level design enables the model to reuse and refine general knowledge while focusing capacity on novel domain specifics when encountering new tasks, achieving efficient adaptation with mitigated forgetting. SEWC operates within the VLP backbone, introducing a targeted parameter regularization technique. SEWC dynamically identifies and consolidates critical backbone parameters exhibiting enduring importance across previously learned tasks, while allowing less crucial parameters to adapt freely to new information. This selective rigidity protects consolidated knowledge from catastrophic forgetting. MAP provides fine-grained, multi-aspect guidance at the input level, while SEWC enforces targeted stability within the model's parameters. This synergistic approach achieves an optimal balance between adaptation and stability without storing historical data, offering a practical, privacy-compliant pathway towards robust, lifelong PAD systems.

Our contributions are summarized as follows:
\begin{itemize}
    \item We propose SVLP-IL, a novel rehearsal-free incremental learning framework for PAD that successfully leverages VLP models for an optimal balance between stability and plasticity.
    \item We introduce MAP, a hierarchical  prompting strategy enabling nuanced, domain-specific adaptation for incremental PAD within VLP models. MAP enhances specialization while inherently mitigating forgetting through explicit knowledge separation and semantic anchoring.
    
    \item We demonstrate that SEWC provides essential complementary stability for the prompt-based RF-IL by dynamically consolidating critical cross-task parameters, offering advantages over backbone freezing, traditional EWC, and other RF-IL strategies, all without rehearsal data.
    \item Extensive experiments show that SVLP-IL significantly mitigates forgetting and achieves superior performance on challenging multi-domain PAD benchmarks, exhibiting exceptional stability, especially over long task sequences.
\end{itemize}

% \begin{figure}[!t]
%     \centering
%     \includegraphics[width=\linewidth]{increPic3.png}
%     \caption{Domain Adaptation (DA) and Domain Generalization (DG) assume simultaneous access to data from multiple domains. Rehearsal-based incremental learning assumes that part of the data from previous domains can be used. In contrast, rehearsal-free incremental learning can only access data from one domain at a time, and data from previous domains is not accessible.}
%     \label{Figure 3:}
% \end{figure}
\begin{figure}[!t]
    \centering
    \includegraphics[width=\linewidth]{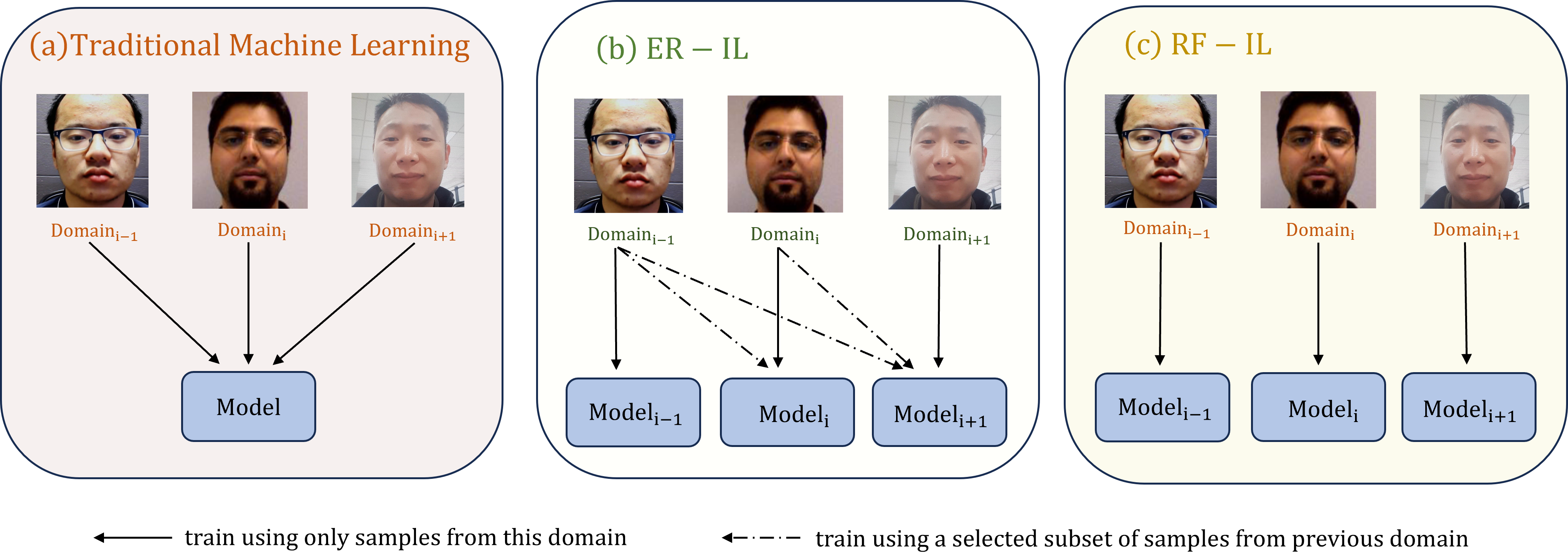}
    \caption{Comparison of training data access assumptions among Traditional Machine Learning, Experience Replay Incremental Learning (ER-IL), and Rehearsal-Free Incremental Learning (RF-IL). (a) Traditional Machine Learning assumes simultaneous access to data from multiple domains. (b) ER-IL assumes that part of the data from previous domains can be used. (c) In contrast, RF-IL can only access data from one domain at a time, and data from previous domains is not accessible.}
    \label{Figure 3:}
\end{figure}

\section{Related Work}
\label{gen_inst}

\subsection{Face Presentation Attack Detection}
% PAD aims to secure FR systems against spoofing attempts. While deep learning has advanced PAD~\cite{Yu2023}, a critical challenge remains: domain shifts due to varying sensors, environments, and continuously evolving attack types severely degrade performance~\cite{li2018}. Traditional Domain Generalization (DG)\cite{jia2020single, wang2022domain} and Adaptation (DA)\cite{li2018, jia2021unified} methods offer partial solutions but typically assume multi-source or target data access. These assumptions conflict with the sequential, ever-changing reality of PAD deployments, where systems must learn incrementally from new data streams, necessitating approaches beyond standard DG/DA.

Face Presentation Attack Detection (PAD) is a crucial security mechanism that safeguards Face Recognition (FR) systems from spoofing attempts using artifacts like printed photos, video replays, or 3D masks~\cite{Yu2023, Ramach2017}. Early PAD methods leveraged handcrafted features, including texture descriptors \cite{chingovska2012effectiveness} and motion cues\cite{kollreider2007real}, to distinguish between live faces and presentation attacks. Although effective in controlled conditions, these methods often struggled with the diversity of attack types and environmental variations.

The emergence of deep learning, particularly Convolutional Neural Networks (CNNs), significantly advanced the field of PAD~\cite{Yu2023, yang2019face}. Deep models learn discriminative features directly from data, enabling them to capture subtle, fine-grained cues associated with spoofing artifacts beyond simple binary classification. Common strategies include the analysis of texture inconsistencies, screen moiré patterns, printing defects, reflections, and inconsistencies in depth or physiological signals~\cite{liu2018learning, Hu2021}. To encourage models to learn intrinsic spoofing cues rather than superficial correlations, researchers have incorporated auxiliary supervision through additional training objectives, such as predicting pixel-wise pseudo-depth maps~\cite{yu2021revisiting} and reflection maps~\cite{Kim2019}.

Despite these advances, the domain shift problem remains a major  challenge~\cite{li2018unsupervised}. Models trained on a specific source domain frequently suffer significant performance degradation when deployed in a new target domain with different sensors, lighting conditions, user demographics, or unseen attack types. To address this issue, researchers have explored Domain Generalization (DG) and Domain Adaptation (DA) techniques. DG methods~\cite{jia2020single, wang2022domain, shao2019multi} typically aim to learn domain-invariant features from multiple diverse source domains, hoping to generalize well to unknown target domains without accessing target data during training. Typical techniques often involve domain randomization, adversarial learning for feature invariance, or meta-learning strategies. More recently, advanced perspectives such as causal intervention for identity debiasing~\cite{long2024dual}, generative reconstruction for anomalous cue mining~\cite{long2025generalized}, and confidence-aware learning for reliability assessment~\cite{long2025confidence} have been proposed to further enhance robustness against unseen scenarios.

In contrast, DA methods~\cite{jia2021unified} typically assume access to target domain data, whether labeled or unlabeled, and align the feature distributions between source and target domains via techniques such as distribution matching or adversarial adaptation.

However, both DG and DA have limitations in real-world PAD scenarios. DG may fail when target domains differ substantially from the source domains used in training. While DA requires prior access to target data and is not designed for sequential adaptation across multiple new domains, its practicality is further constrained by strict privacy considerations. Such regulations impose tight restrictions on the storage and reuse of past data, thereby limiting the feasibility of standard DA in continuous learning scenarios. These operational constraints, i.e., continuous domain evolution and strict data privacy, highlight the need for robust Incremental Learning (IL) solutions tailored for PAD.

% DA typically requires knowledge of and access to the target domain data beforehand, and standard approaches are not inherently designed for sequential adaptation across multiple, continuously arriving new domains where past data cannot be stored due to privacy regulations. This operational reality—characterized by continuous evolution and strict data privacy constraints—underscores the fundamental need for robust Incremental Learning capabilities specifically tailored for the challenges of PAD.

\subsection{Incremental Learning for PAD}
Incremental Learning (IL), also known as Continual Learning, aims to enable models to learn sequentially from new data without catastrophic forgetting~\cite{kirkpatrick2017, wang2024comprehensive, delange2021survey}. This capability is crucial for real-world PAD deployments where attack types and environmental conditions evolve continuously. 

A common IL strategy is experience replay, which stores a subset of past data and replays it during new training sessions~\cite{Rostami2021}. Early PAD-specific IL approaches, such as the continual meta-learning framework by Pérez-Cabo et al.~\cite{firstPADIncre2020} and the novel attack learning method by Rostami et al.~\cite{Rostami2021}, successfully employed replay. Despite their effectiveness, rehearsal-based methods are often infeasible for PAD due to privacy restrictions that prevent the storage and reuse of facial data~\cite{cai2023, wang2024multi}. This limitation has motivated the development of Rehearsal-Free Incremental Learning (RF-IL) for practical PAD systems.

RF-IL methods primarily fall into two categories: regularization-based  and parameter isolation-based approaches. Regularization-based methods introduce penalty terms in the loss function to prevent significant updates to parameters important for previous tasks. Notable examples include Elastic Weight Consolidation (EWC)\cite{kirkpatrick2017}, Synaptic Intelligence (SI)\cite{zenke2017continual}, and Learning without Forgetting (LwF)\cite{li2017learning}. A key challenge is balancing stability, which entails preserving old knowledge, with plasticity, which allows for acquiring new knowledge. Overly strict regularization hinders adaptation, while insufficient regularization leads to forgetting. Furthermore, standard regularization signals may be inadequate for the fine-grained nature of PAD tasks. For instance, Guo et al.\cite{guo2022multi} argued that typical logits or activation maps fail to capture nuanced spoof regions, proposing Spoof Region Estimator (SRE) as a more effective data-driven regularization signal within the FAS-wrapper framework. Similarly, Cai et al.~\cite{cai2023} introduced Proxy Prototype Contrastive Regularization (PPCR), to approximate past class centroids without storing data. Our proposed Selective Elastic Weight Consolidation (SEWC) extends this idea, aiming for targeted regularization to preserve flexibility.

Alternatively, parameter isolation methods allocate dedicated model parameters to different tasks or domains to prevent interference~\cite{rebuffi2018efficient}. These methods may involve growing the network, employing task-specific experts, or using masking strategies. While effective against forgetting, they may face scalability issues as the number of parameters grows, limited knowledge transfer across tasks, and difficulties in selecting appropriate parameters during inference. Wang et al.\cite{wang2024multi} addressed this for multi-domain incremental PAD with an Adaptive Domain-specific Experts (ADE) framework, which incorporates an Instance-wise Router (IwR) for test-time expert selection. Parameter-Efficient Fine-Tuning (PEFT) techniques, such as using adapters~\cite{cai2023} or prompts~\cite{wang2022_sprompts, L2P_ref}, can also be viewed as parameter isolation methods, as they train small, task-specific components while keeping the backbone network largely frozen.

\subsection{Prompt Learning for Vision-Language Pre-trained Models}
Prompt Learning has emerged as a prominent parameter-efficient fine-tuning (PEFT) paradigm for adapting Vision-Language Pre-trained (VLP) models like CLIP \cite{radford2021}. Instead of full fine-tuning, it learns task-specific vectors to condition the model\cite{jia2022visual, lester2021power}. This approach has been widely explored in both domain generalization and incremental learning.

In domain generalization for Face Anti-Spoofing (FAS), prompt designs have grown increasingly sophisticated. Methods like Fine-Grained Prompt Learning (FGPL)~\cite{fineGrained} decompose prompts into domain-agnostic and domain-specific components to learn more robust representations. In Incremental Learning (IL), key strategies include managing a dynamic prompt pool~\cite{L2P_ref}, decomposing prompts into shared and task-specific parts~\cite{feng2024cp}, or learning independent prompts for each task with a frozen backbone~\cite{wang2022_sprompts}.

However, directly applying these strategies to  rehearsal-free IL for PAD reveals a stability-plasticity dilemma. Detecting subtle spoofing artifacts requires the model's core visual features to adapt, yet a fully frozen backbone~\cite{wang2022_sprompts} inhibits plasticity. Conversely, updating the backbone without constraints leads to catastrophic forgetting of both foundational VLP knowledge and previously learned spoofing cues.

Our SVLP-IL framework addresses this challenge by combining a Multi-Aspect Prompting (MAP) strategy with Selective Elastic Weight Consolidation (SEWC). Inspired by decomposition methods like FGPL~\cite{fineGrained}, MAP isolates domain-specific knowledge at the prompt level. SEWC then enables controlled adaptation of the VLP backbone, allowing it to learn new visual cues while protecting essential prior knowledge. This combination balances adaptability and stability, overcoming the limitations of methods that either overly restrict or insufficiently regulate backbone updates.

\section{Methodology}
\label{headings}

\subsection{Framework Overview}
In the domain-incremental learning (IL) setting for PAD, a model sequentially learns from distinct domains $\mathcal{D}_1, \dots, \mathcal{D}_T$ without retaining samples from earlier domains. The challenge is to sustain high performance across all encountered domains $\mathcal{D}_{1..T}$ while remaining generalizable to unseen domains, despite the constant evolution of attack methods and the wide variety of real-world scenarios, such as different cameras and lighting conditions.

\begin{figure*}[htbp]
    \centering
    \includegraphics[width=\textwidth]{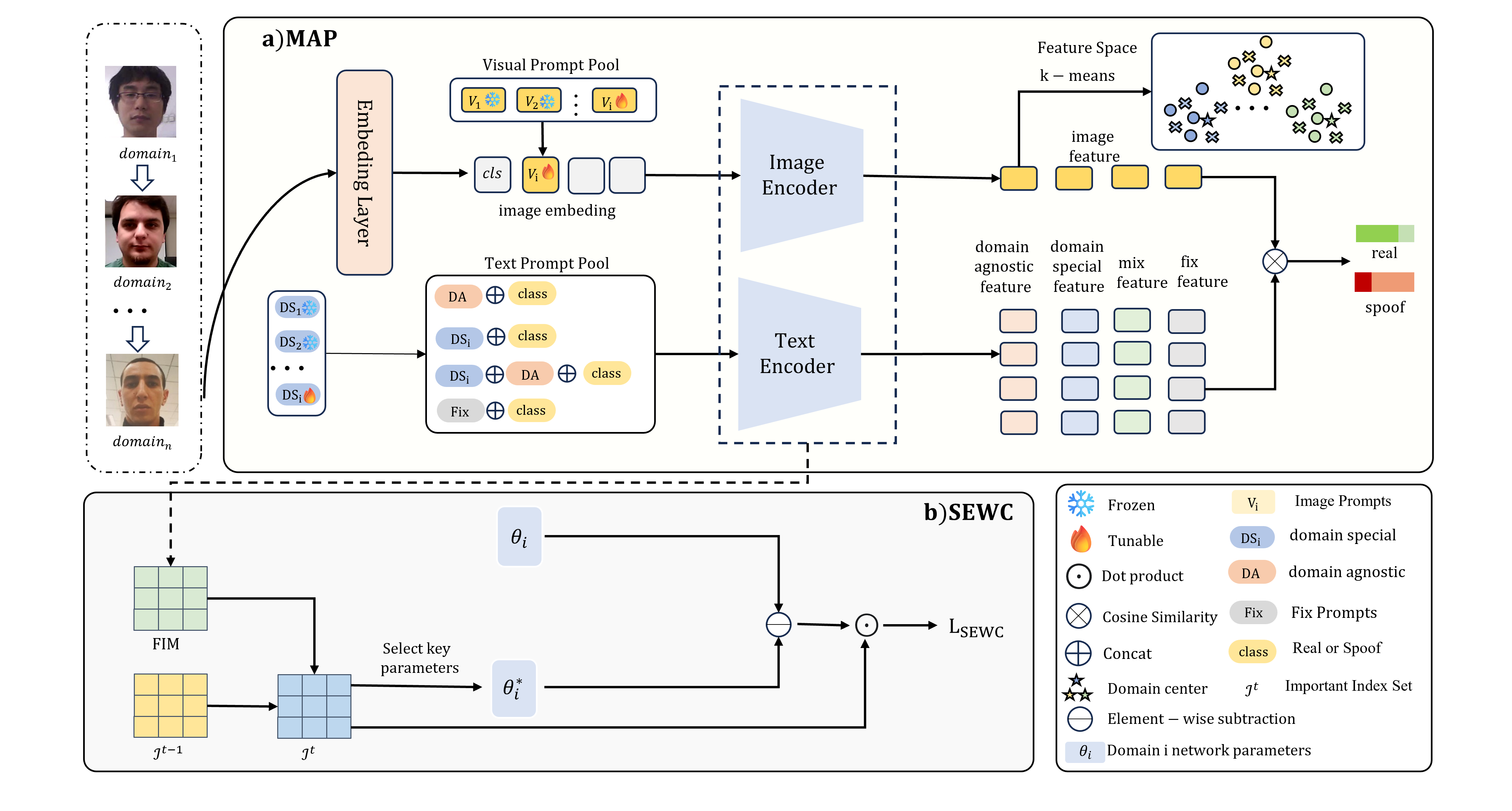}
    \caption{
       Overview of the SVLP-IL framework. The framework learns incrementally by balancing adaptation and stability. MAP adapts the model to new domains by learning a structured set of visual and textual prompts that capture both domain-specific and universal spoofing cues. Concurrently, SEWC preserves existing knowledge by protecting critical parameters in the shared backbone from being overwritten. During inference, a lightweight prototype-based router selects the appropriate prompts for a given test image.
      }
    \label{Figure 1:}
\end{figure*}

Our proposed framework, SVLP-IL, is built upon CLIP~\cite{radford2021}, as shown in Fig.~\ref{Figure 1:}. It leverages CLIP’s pre-trained image and text encoders that project both modalities into a shared embedding space. The central idea of SVLP-IL is to steer the adaptation of the vision–language pre-trained model through two synergistic mechanisms that jointly balance adaptability to new domains and retention of previously acquired knowledge, all under the rehearsal-free constraint.

The first mechanism, Multi-Aspect Prompting (MAP), enables efficient domain-level adaptation by injecting learnable prompts into both the visual and textual pathways. These prompts capture domain-specific and shared semantic cues, guiding the model to adjust representations with minimal parameter updates.  
The second mechanism, Selective Elastic Weight Consolidation (SEWC), complements MAP by stabilizing the shared backbone parameters. It identifies the most influential weights accumulated over past domains and selectively regularizes them, preventing catastrophic forgetting while maintaining sufficient flexibility for learning new domain cues.

During inference, we must select the most relevant prompts for test images that may come from unseen domains, without retaining historical data. We therefore build a lightweight prototype bank: for each domain, we pre-compute \(k\)-means prototypes in the shared feature space. A test image is embedded and assigned to its nearest prototype, which activates the corresponding domain-specific prompts.  This privacy-preserving, memory-efficient routing approximates domain recognition. The subsequent sections detail Multi-Aspect Prompting (MAP) and Selective Elastic Weight Consolidation (SEWC).

\subsection{Multi-Aspect Prompting (MAP)}

MAP comprises two coordinated prompting modules: a visual prompting module that adapts the image pathway with domain-specific tokens, and a textual prompting module that forms four prompt families—domain-agnostic (DA), domain-specific (DS), mixed (Mix), and fixed (Fixed)—to provide complementary semantic guidance. This design isolates domain idiosyncrasies at the prompt level while keeping a stable, shared semantic foundation across domains, thereby reducing interference across domains and easing knowledge transfer.

\paragraph{Visual prompting}
For each domain $t$, we maintain learnable visual prompt tokens $D_V(t)\in\mathbb{R}^{L_v\times C}$, where $L_v$ is the number of visual prompt tokens and $C$ is the token embedding dimension. During training on domain $t$, only $D_V(t)$ is updated and previously learned $\{D_V(j)\}_{j<t}$ are frozen. Let $X_{\text{cls}}$ denote the [CLS] token and $X_{\text{img}}(t)$ the image patch embeddings. We insert $D_V(t)$ after [CLS] to form the input sequence to the image encoder:
\begin{equation}
P_v(t)=\operatorname{Concat}\!\big(X_{\text{cls}},\,D_V(t),\,X_{\text{img}}(t)\big).
\label{eq:visual_prompt_concat}
\end{equation}
Passing $P_v(t)$ through the image encoder yields the visual feature
\begin{equation}
f^{\text{img}} \,=\, \operatorname{ImageEnc}\!\big(P_v(t)\big)\ \in\ \mathbb{R}^{C_{\text{out}}},
\label{eq:fI_from_imageenc}
\end{equation}
where $C_{\text{out}}$ is the output embedding dimension of the encoder.

\paragraph{Textual prompting}
We use two types of learnable textual contexts: a continually updated domain-agnostic context $D_A\in\mathbb{R}^{N_{\text{ctx}}\times C}$ shared by all domains, and a domain-specific context $D_S(t)\in\mathbb{R}^{N_{\text{ctx}}\times C}$ updated only when training on domain $t$. Let $E_{\text{cls}}\in\mathbb{R}^{N_{\text{cls}}\times C}$ denote the frozen class-name embeddings for the labels ``real'' and ``spoof''. At step $t$, we construct three dynamic prompt families:
\begin{align}
P_{\text{da}}       &= \operatorname{Concat}\!\big([\text{sos}],\, D_A,\,       E_{\text{cls}},\, [\text{eos}]\big),\\
P_{\text{ds}}(t)    &= \operatorname{Concat}\!\big([\text{sos}],\, D_S(t),\,    E_{\text{cls}},\, [\text{eos}]\big),\\
P_{\text{mix}}(t)   &= \operatorname{Concat}\!\big([\text{sos}],\, D_A,\, D_S(t),\, E_{\text{cls}},\, [\text{eos}]\big),
\end{align}
Finally, we include a fixed prompt family whose texts are immutable natural-language sentences associated with the class names. We denote this collection by $P_{\text{fixed}}$, which remains constant across all domains. The fixed natural-language sentences associated with the class names are fed to the text encoder directly, without any learned context.

Feeding these prompts to the text encoder produces per-class text features for each family \(k \in K\), where \(K = \{\text{da}, \text{ds}, \text{mix}, \text{fixed}\}\):
\begin{equation}
T_k(t)=\operatorname{TextEnc}\!\big(P_k(t)\big)\in\mathbb{R}^{N_{\text{cls}}\times C_{\text{out}}}.
\end{equation}
Let $\{f^{\text{txt}}_{k,c}(t)\}_{c=1}^{N_{\text{cls}}}$ denote the per-class text feature vectors obtained from $T_k(t)$, with $f^{\text{txt}}_{k,c}(t)\in\mathbb{R}^{C_{\text{out}}}$ corresponding to class $c$.

% \paragraph{Roles of the prompt families}
% DA captures cross-domain, transferable semantics accumulated across all domains. DS focuses on domain-$t$ idiosyncrasies. Mix composes both to provide holistic semantics that couple general cues with domain specifics. Fixed offers a stable natural-language anchor that leverages CLIP’s zero-shot prior and regularizes drift in the text pathway. This division of responsibilities allows domain specialization without overwriting previously learned prompts, while DA and Mix promote forward transfer. 

\paragraph{Synergy of visual and textual prompts}
Our framework explicitly decomposes the domain adaptation process into visual and textual aspects. 
The visual prompts operate at the input stage to absorb low-level domain shifts, such as variations in image resolution, lighting conditions, or sensor-specific noise. By handling these visual idiosyncrasies early, they relieve the burden on the textual pathway, preventing the semantic prompts from overfitting to non-semantic visual artifacts.
Complementing this, the textual prompt families capture high-level semantic variations:
DA consolidates cross-domain, transferable spoofing cues shared by all tasks. 
DS specializes in handling domain-$t$ specific semantic characteristics. 
Mix bridges the two, providing a holistic representation that couples general knowledge with domain specifics. 
Finally, Fixed serves as a stable anchor, leveraging CLIP's zero-shot prior to regularize the text pathway against catastrophic forgetting.

\paragraph{Class-wise logits and MAP loss}
We perform L2 normalization on the visual feature and each class-wise text vector: $\widehat{f}^{\text{img}}=f^{\text{img}}/\|f^{\text{img}}\|_2$ and $\widehat{f}^{\text{txt}}_{k,c}(t)=f^{\text{txt}}_{k,c}(t)/\|f^{\text{txt}}_{k,c}(t)\|_2$. Given a learnable temperature $\tau$, the class-wise logit for family $k$ is
\begin{equation}
\big[\operatorname{logit}_k(t)\big]_c=\tau\,\big\langle \widehat{f}^{\text{img}},\,\widehat{f}^{\text{txt}}_{k,c}(t)\big\rangle,\quad \operatorname{logit}_k(t)\in\mathbb{R}^{N_{\text{cls}}}.
\label{eq:logit_family}
\end{equation}
Here $\langle\cdot,\cdot\rangle$ denotes the standard Euclidean inner product.

With $K=\{\text{da},\,\text{ds},\,\text{mix},\,\text{fixed}\}$ and let
\[
\alpha^{(t)}=\big(\alpha^{(t)}_{\text{da}},\,\alpha^{(t)}_{\text{ds}},\,\alpha^{(t)}_{\text{mix}},\,\alpha^{(t)}_{\text{fix}}\big)^\top\in\mathbb{R}^{|K|},
\]
which denotes the domain-specific logits over prompt families at step $t$. During training on domain $t$, $\alpha^{(t)}$ is optimized jointly with MAP parameters, while previously learned $\{\alpha^{(j)}\}_{j<t}$ are frozen and stored. We convert $\alpha^{(t)}$ to aggregation weights by a softmax:
\begin{equation}
w_k^{(t)}=\operatorname{softmax}_k(\alpha^{(t)}),\qquad k\in K.
\end{equation}

For clarity, we first aggregate the per-family logits into a single vector:
\begin{equation}
\operatorname{logit}_{\text{agg}}(t)\;=\;\sum_{k\in K} w_k^{(t)}\,\operatorname{logit}_k(t)\;\in\mathbb{R}^{N_{\text{cls}}}.
\label{eq:agg_logit}
\end{equation}

The MAP objective then minimizes a single cross-entropy on the aggregated logits:
\begin{equation}
L_{\text{MAP}}(t)\;=\;\mathcal{L}_{\text{CE}}\!\big(\operatorname{logit}_{\text{agg}}(t),\,y\big).
\label{eq:map_loss}
\end{equation}

\subsection{Selective Elastic Weight Consolidation (SEWC)}
\label{subsec:sewc}
Regularization-based continual learning methods such as
Elastic Weight Consolidation (EWC)~\cite{kirkpatrick2017}
stabilize previously learned knowledge by penalizing departures from past optima.
This quadratic penalty is not ad hoc: it corresponds to the negative
log of a Gaussian approximation to the previous posterior. In our rehearsal-free PAD
setting with a large VLP backbone, penalizing all network parameters across a
long sequence over-regularizes and hinders adaptation. We therefore revisit EWC
from a Bayesian perspective and adopt a selective variant that retains per-domain optima and applies the quadratic only to consistently important network parameters.

\paragraph{Bayesian view and notation}
Let $\theta$ denote the vector of trainable VLP backbone parameters that are subject to consolidation, which comprise the image encoder and the text encoder.
We write $\theta_i$ for the $i$-th coordinate.
Let $\theta^{*(t)}$ denote the parameter vector obtained at the end of training on domain $t$, and let $\theta^{*(t)}_i$ denote its $i$-th coordinate.
Let $D_{1..t}$ be the data from domains $1$ through $t$. By Bayes' rule,
\begin{equation}
p(\theta \mid D_{1..t}) \propto p(D_t \mid \theta)\, p(\theta \mid D_{1..t-1}).
\label{eq:bayes}
\end{equation}
Taking the negative log and dropping $\theta$-independent constants
leads to the maximum-a-posteriori objective:
\begin{equation}
L(\theta) \approx L_{\text{task}}^{(t)}(\theta) - \log p(\theta \mid D_{1..t-1}).
\label{eq:map_objective_general}
\end{equation}
where $L_{\text{task}}^{(t)}(\theta)$ is the negative log-likelihood on $D_t$.

\paragraph{Laplace approximation and diagonal Fisher}
We approximate the intractable prior $p(\theta \mid D_{1..t-1})$
using a second-order Laplace expansion around the previous
optimum $\theta^{*(t-1)}$\cite{huszar2017ewc}:

\begin{algorithm}[H] 
\caption{SVLP-IL training with SEWC}
\label{alg:budgeted_sewc}
\begin{algorithmic}[1]
    \Require Datasets $\{\mathcal{D}_t\}_{t=1}^{T}$; initial parameters $\theta^{(0)}$; percentile $p\in(0,1]$
    \Ensure Final parameters $\theta^{(T)}$
    
    \State $\mathcal{I}^{(0)}\gets\varnothing$ \Comment{Important Index Set}
    \State $\mathcal{S}\gets \mathrm{IndexSet}(\theta_{\text{network}})$ \Comment{indices of penalizable params}
    \State $\mathcal{M}\gets\varnothing$ \Comment{cache of historical $(F_j,\;\theta^{*(j)})$ }
    
    \For{$t=1$ \textbf{to} $T$}
        \ForAll{mini-batches from $\mathcal{D}_t$}
            \State $L_{\mathrm{MAP}} \gets \textsc{GetMapLoss}(\theta,\mathcal{D}_t)$ \Comment{Eq.~\eqref{eq:map_loss}}
            \If{$t>1$ \textbf{and} $\mathcal{I}^{(t-1)}\neq\varnothing$}
                \State $L_{\mathrm{SEWC}} \gets \textsc{GetSEWCLoss}(\theta;\mathcal{I}^{(t-1)},\mathcal{M})$
\Statex \hspace*{\algorithmicindent}$\triangleright$~Eq.~\eqref{eq:sewc_budget}

            \Else
                \State $L_{\mathrm{SEWC}} \gets 0$
            \EndIf
            \State $L_{\text{total}} \gets L_{\mathrm{MAP}} + L_{\mathrm{SEWC}}$
            \State $\theta \gets \textsc{Update}(\theta, L_{\text{total}})$
        \EndFor

        \State $\theta^{*(t)} \gets \theta$ \Comment{snapshot of domain-$t$ optimum}
        \State $F_t \gets \textsc{EstimateFisher}\big(\theta^{*(t)}, \mathcal{D}_t\big)$ \Comment{Eq.~\eqref{eq:fisher_def}}
       
        \State $\tau^{(t)} \gets \operatorname{Quantile}\big((F_t)_{\mathcal{S}},\,1-p\big)$ \Comment{Eq.~\eqref{eq:budget_set}}
        \State $\mathcal{J}^{(t)} \gets \{\,i\in \mathcal{S}:\,(F_t)_i \ge \tau^{(t)}\,\}$ \Comment{Eq.~\eqref{eq:budget_set}}
        \State $\mathcal{I}^{(t)} \gets \mathcal{I}^{(t-1)} \cup \mathcal{J}^{(t)}$ \Comment{Eq.~\eqref{eq:getmask}}
        \State $\mathcal{M} \gets \mathcal{M} \cup \{(F_t,\theta^{*(t)})\}$ \Comment{Eq.~\eqref{eq:sewc_budget}}
        % \Statex \Comment{\emph{Optional: keep only last $K$ domains in $\mathcal{M}$ to bound memory.}}
    \EndFor
\end{algorithmic}
\end{algorithm}

\begin{align}
\log p(\theta \mid D_{1..t-1}) \approx\ 
& \log p(\theta^{*(t-1)} \mid D_{1..t-1}) \notag\\
& - \tfrac{1}{2} (\theta - \theta^{*(t-1)})^\top H_{t-1} (\theta - \theta^{*(t-1)}).
\end{align}
with $H_{t-1}$ denoting the Hessian of the negative log-posterior at $\theta^{*(t-1)}$.
Following common practice, we approximate \(H_{t-1}\) with a diagonal Fisher information matrix. 
At the end of each domain we estimate the diagonal FIM using only the current domain data. 
Concretely, for the previous domain \(t-1\):

\begin{equation}
F_{t-1,i} \;\approx\; \frac{1}{|D_{t-1}|} \sum_{(x,y)\in D_{t-1}}
\left(\frac{\partial L_{\text{task}}^{(t-1)}(x,y;\,\theta^{*(t-1)})}{\partial \theta_i}\right)^2.
\label{eq:fisher_def}
\end{equation}
By incorporating this approximation into the maximum-a-posteriori objective for the new task $t$, we arrive at the classic EWC loss~\cite{kirkpatrick2017}. The negative log-prior becomes a quadratic penalty term that regularizes parameters based on their importance to the previous task:
\begin{equation}
L_{\text{EWC}}^{(t)} = L_{\text{task}}^{(t)}(\theta) + \frac{1}{2}\sum_i F_{t-1,i} \, (\theta_i - \theta_{i}^{*(t-1)})^2.
\label{eq:classic_ewc}
\end{equation}
The FIM $F_{t-1}$ estimates the importance of each parameter $\theta_i$ for tasks $1..t-1$.

\paragraph{Multi-center prior}
We approximate the accumulated prior $p(\theta\mid D_{1..t-1})$ as a product of Gaussian factors, one for each past domain $j$, obtained by a Laplace expansion around $\theta^{*(j)}$ with diagonal precision $F_j$. This leads to the quadratic
\begin{equation}
L_{\text{prior}}^{(t)} \;=\; \frac{1}{2}\sum_{j=1}^{t-1}\sum_{i} F_{j,i}\,\bigl(\theta_i-\theta^{*(j)}_i\bigr)^2.
\label{eq:online_ewc_dense}
\end{equation}
The penalty anchors $\theta$ to every past optimum in coordinates with large Fisher values, which preserves previously learned knowledge while keeping low-importance directions flexible for adapting to $D_t$.

\paragraph{Selective consolidation}

Penalizing all coordinates in Eq.~\eqref{eq:online_ewc_dense} can still
over-constrain the model as $t$ grows. SEWC therefore applies the
quadratic only to a subset selected by a union-of-top-$p$ rule.
Let $S$ denote the index set of penalizable backbone parameters and let $M=|S|$.
Let $p\in(0,1]$ denote the fraction to keep per domain. For each historical domain $j$,
we form a per-domain top-$p$ set by quantile thresholding on the diagonal FIM:
\begin{equation}
\begin{aligned}
\tau^{(j)} \;&=\; \operatorname{Quantile}\!\big(F_{j,S},\, 1-p\big), \\
\mathcal{J}^{(j)} \;&=\; \big\{\, i\in S \,:\, F_{j,i}\ge \tau^{(j)} \,\big\}.
\end{aligned}
\label{eq:budget_set}
\end{equation}
Here, $\operatorname{Quantile}(v,q)$ denotes the smallest threshold $\tau$ such that at least a fraction $q$
of the entries in $v$ are less than or equal to $\tau$. We then keep the union over all past domains to obtain an important index set.
\begin{equation}\label{eq:getmask}
\mathcal{I}^{(t-1)} \;=\; \bigcup_{j=1}^{t-1} \mathcal{J}^{(j)}.
\end{equation}
The selective penalty used while optimizing on $D_t$ is
\begin{equation}
L_{\text{SEWC}}{(t)} \;=\; \frac{1}{2}\sum_{i\in\mathcal{I}^{(t-1)}}\;\sum_{j=1}^{t-1}
F_{j,i}\,\bigl(\theta_i - \theta^{*(j)}_i\bigr)^2.
\label{eq:sewc_budget}
\end{equation}
This union-based selection captures parameters  that are repeatedly identified as important across domains and strengthens retention of shared knowledge.

% \paragraph{Fisher estimation and updates}
% After finishing domain $t$, we estimate the diagonal FIM at $\theta^{*(t)}$ as
% \begin{equation}
% F_{t,i} \approx \frac{1}{|D_t|}\sum_{(x,y)\in D_t}
% \left(\frac{\partial L_{\text{task}}^{(t)}(x,y;\theta^{*(t)})}{\partial \theta_i}\right)^2.
% \label{eq:fisher_empirical}
% \end{equation}
% We then compute $\tau^{(t)}$ and $\mathcal{J}^{(t)}$ by Eq.~\eqref{eq:budget_set} and update the
% union set by Eq.~\eqref{eq:getmask}.
% % (Optionally, one may cache only entries with $i\in\mathcal{I}^{(t)}$.)

\subsection{Overall Objective and Inference}
The total training objective for domain $t$ is a combination of the Multi-Aspect Prompting loss and the Selective Elastic Weight Consolidation penalty:
\begin{equation}
L_{\text{total}}(t) = L_{\text{MAP}}(t) + L_{\text{SEWC}}{(t)}.
\label{eq:total_loss}
\end{equation}
This composite loss function works synergistically: MAP facilitates adaptation to new domains by isolating new knowledge in dedicated prompts, which inherently helps prevent overwriting shared knowledge. SEWC complements this by explicitly stabilizing the most critical shared parameters in the VLP backbone, further ensuring the retention of previously learned information.

During inference, the model must select the appropriate domain-specific components for a given test image, which may come from an unseen domain. To accomplish this in a privacy-preserving manner without storing past data, we employ a lightweight routing mechanism based on a pre-computed prototype bank. For each previously learned domain, we run k-means clustering on its training image features ($f^{\text{img}}$) to obtain $k$ representative prototypes. Given a test image, we compute its feature embedding and find the nearest prototype in the entire bank. The domain corresponding to this prototype, $\hat{t}$, is selected. This selection activates the corresponding visual prompt ($D_V(\hat{t})$), domain-specific textual context ($D_S(\hat{t})$), and aggregation weights ($w_k^{(\hat{t})}$). The final prediction is then computed by aggregating the logit families using these selected weights:
\begin{equation}
\operatorname{logit_{agg}}=\sum_{k\in K} w_k^{(\hat t)}\,\operatorname{logit}_k(\hat t).
\label{eq:test_weighted_final}
\end{equation}

\section{Experiment}
\subsection{Datasets}
We evaluate our method using nine publicly available PAD datasets, selected to encompass a wide range of attack types, sensors, and acquisition conditions for robust incremental learning assessment. These datasets include Idiap REPLAY-ATTACK~\cite{chingovska2012effectiveness}, CASIA-FASD~\cite{zhang2012face}, MSU-MFSD~\cite{wen2015face}, HKBU-MARs-V1+~\cite{liu20163d}, OULU-NPU~\cite{boulkenafet2017oulu}, WFFD~\cite{jia2019database}, ROSE-YOUTU~\cite{li2018unsupervised}, SiW~\cite{liu2018learning}, and the large-scale CelebA-Spoof dataset~\cite{zhang2020celeba}.

\subsection{Implementation Details}

% Our implementation is based on PyTorch~\cite{paszke2019pytorch} using the pre-trained CLIP ViT-B/16 model~\cite{radford2021} as the backbone of the vision-language model. All input images are resized to \(224 \times 224\). We train with a batch size of 8 for 100 iterations per domain, using a fixed random seed. Both the learning rate and the weight decay are set to \(1 \times 10^{-5}\), with the Adam optimizer. For Multi-Aspect Prompting (MAP), the visual prompts have a length \(L_v = 16\). The hierarchical text prompts employ domain-agnostic and domain-specific context vectors, each of length \(N_{ctx} = 16\).  The fixed text prompt is “This is a photo of {real/spoof} face.” For Selective Elastic Weight Consolidation (SEWC), we protect the top 50\% of parameters via quantile thresholding. During inference, we perform \(k\)-means clustering (\(k=5\)) on training features to obtain domain centroids and assign prompts to test samples by using \(\ell_2\) distance.

Our implementation is based on PyTorch~\cite{paszke2019pytorch} using the pre-trained CLIP ViT-B/16 model~\cite{radford2021} as the backbone of the vision-language pre-trained model. All input images are resized to \(224 \times 224\). We train with a batch size of 8 for 100 iterations per domain, using a fixed random seed. Both the learning rate and the weight decay are set to \(1 \times 10^{-5}\), with the Adam optimizer. For Multi-Aspect Prompting (MAP), the visual prompts have a length \(L_v = 16\). The hierarchical text prompts employ domain-agnostic and domain-specific context vectors, each of length \(N_{ctx} = 16\).  The fixed text prompt is “This is a photo of {real/spoof} face.” For Selective Elastic Weight Consolidation (SEWC), we protect the top 50\% of parameters via quantile thresholding. During inference, for each seen domain we run k-means with $k=5$ prototypes per domain on its training features. A test sample is routed to the domain whose prototype is closest in $\ell_2$, and the corresponding prompts are applied.

\subsection{Evaluation Metrics}

% To assess the performance of our approach and baseline methods, we employ widely recognized metrics in the PAD field. Following standard practices~\cite{jia2020single}, we utilize the Half Total Error Rate (HTER), which balances false acceptance and false rejection rates, and the Area Under the Receiver Operating Characteristic Curve (AUC), reflecting the overall discriminative ability across different thresholds. Lower HTER and higher AUC indicate better performance.

% To specifically quantify the model's ability to retain performance on previously learned domains relative to a multi-domain model , we adopt a metric proposed by~\cite{wang2024multi}, denoted as \(\Delta m\%\). This metric calculates the average normalized decrease in HTER on each task \(t\) for a given model \(q\) compared to this multi-domain reference \(b\), and is computed as:
We adopt standard PAD metrics, i.e., Half Total Error Rate (HTER) and Area Under the ROC Curve (AUC), to evaluate performance. Lower HTER and higher AUC indicate better detection accuracy. To specifically measure forgetting in incremental learning, we use the \(\Delta m\%\) metric~\cite{wang2024multi}, defined as:

\begin{equation}
\Delta m\% = \frac{1}{T} \sum_{t=1}^{T} \frac{\text{HTER}_{q,t} - \text{HTER}_{b,t}}{1 - \text{HTER}_{b,t}}
\label{eq:delta_m}
\end{equation}

% \noindent where \(\text{HTER}_{b,t}\) and \(\text{HTER}_{q,t}\) are the HTER scores of the  multi-domain reference \(b\) and the incremental model \(q\) on task \(t\), respectively. A lower \(\Delta m\%\) value indicates better retention of prior knowledge.

\noindent where \( T \) is the number of learned tasks. \(\text{HTER}_{q,t}\) denotes the HTER of the current incremental model evaluated on historical task \( t \). \(\text{HTER}_{b,t}\) represents the reference HTER obtained by Joint Training (JT) on task \( t \), serving as the performance upper bound. The denominator normalizes the error gap, and a lower \( \Delta m\% \) value indicates that the incremental model effectively retains prior knowledge, performing closer to the joint training baseline.

\subsection{Results of Domain Incremental Settings}

We compare against two key baselines: Fine-tuning (FT), which updates all parameters sequentially without anti-forgetting mechanisms, and Joint Training (JT), which trains on all domains simultaneously and serves as an upper bound. We evaluate on Protocol-1,  where models learn incrementally from MSU-MFSD (M), CASIA-FASD (C), Idiap REPLAY-ATTACK (I), and OULU-NPU (O).

As shown in Table~\ref{tab:exp1}, SVLP-IL demonstrates strong stability-plasticity trade-off. After learning the second domain (C), it maintains an HTER of only 0.43\% on the first domain (M), significantly outperforming FT and other methods. This trend continues throughout incremental steps, i.e., after learning the fourth domain (O), SVLP-IL achieves HTERs of 2.38\%, 2.14\%, and 1.28\% on M, C, and I, respectively, indicating minimal forgetting.

While SVLP-IL occasionally attains slightly higher HTER on the most recent task compared to FT , such as 2.59\% versus 0.00\% on domain O, it excels in overall knowledge retention, as reflected in its lowest $\Delta m\%$ value across all steps. This result confirms that SVLP-IL effectively balances adaptation to new tasks with preservation of past knowledge, outperforming methods that either forget aggressively or overspecialize on the current domain.

\begin{table*}[htbp]
\centering
% \caption{Results (HTER in \% and $\Delta m\%$) for the first multi-domain incremental learning scenario (MSU~$\rightarrow$~CASIA~$\rightarrow$~Idiap~$\rightarrow$~OULU sequence). Performance is evaluated cumulatively after each domain is learned. Lower values are better.
% }
\caption{Multi-domain incremental learning performance in terms of HTER and $\Delta m\%$ on the MSU-MFSD~$\rightarrow$~CASIA-MFSD~$\rightarrow$~Idiap Replay-Attack~$\rightarrow$~OULU-NPU sequence. Performance is evaluated cumulatively after each domain is learned. Lower values are better
}
\label{tab:exp1}
% \vspace{1mm}
\small
\resizebox{\textwidth}{!}{ % 正确位置：包裹整个 tabular 环节
\begin{tabular}{lcccccccccccc}
\toprule
\multirow{2}{*}{\textbf{Methods}} & \multicolumn{3}{c}{\textbf{Step 2 (M→C)}} & \multicolumn{4}{c}{\textbf{Step 3 (M→C→I)}} & \multicolumn{5}{c}{\textbf{Step 4 (M→C→I→O)}} \\
\cmidrule(r){2-4} \cmidrule(r){5-8} \cmidrule(r){9-13}
 & M↓ & C↓ & $\Delta m\%$↓ & M↓ & C↓ & I↓ & $\Delta m\%$↓ & M↓ & C↓ & I↓ & O↓ & $\Delta m\%$↓ \\
\midrule
JT & 0.43 & 0.00 & - & 0.81 & 0.00 & 0.00 & - & 2.82 & 0.00 & 0.13 & 1.12 & - \\
FT & 6.07 & 0.00 & 2.83 & 6.07 & 1.17 & 0.00 & 2.16 & 9.76 & 2.14 & 1.03 & 0.00 & 2.26 \\
\midrule
LwF~\cite{li2017learning} & 20.06 & 0.02 & 9.87 & 24.17 & 23.56 & \textbf{0.00} & 15.70 & 12.13 & 23.54 & 27.50 & 0.54 & 14.99 \\
DyTox~\cite{douillard2022dytox} & 9.24 & 1.00 & 4.92 & 21.11 & 3.25 & 0.70 & 8.14 & 19.93 & 4.40 & 24.14 & 12.14 & 14.30 \\
L2P~\cite{L2P_ref} & 8.49 & 0.15 & 4.12 & 16.36 & 4.13 & 0.84 & 6.89 & 17.36 & 11.59 & 15.98 & 0.92 & 10.56 \\
S-iPr.~\cite{wang2022_sprompts} & 3.20 & \textbf{0.00} & 1.39 & 6.20 & \textbf{1.13} & 2.09 & 2.89 & 11.35 & 2.38 & 2.09 & 3.01 & 3.76 \\
MDIL-PAD~\cite{wang2024multi} & 6.52 & \textbf{0.00} & 3.05 & 5.41 & 1.57 & \textbf{0.00} & 2.07 & 10.62 & 2.54 & 1.44 & \textbf{0.31} & 2.77 \\
SD-LoRA~\cite{wu2025sdlora} & 8.89 & 0.19 & 4.34 & 12.15 & 0.00 & 1.41 & 4.28 & 12.58 & 0.19 & 1.41 & 6.17 & 4.16 \\
KA-Prompt~\cite{xu2025kaprompt} & 6.51 & \textbf{0.00} & 3.05 & 8.89 & 0.00 & 3.60 & 3.92 & 12.15 & \textbf{0.00} & 2.69 & 2.24 & 3.32 \\

\textbf{Ours} & \textbf{0.43} & 2.33 & \textbf{1.17} & \textbf{0.43} & 1.17 & 1.03 & \textbf{0.61} & \textbf{2.38} & 2.14 & \textbf{1.28} & 2.59 & \textbf{1.08} \\
\bottomrule
\end{tabular}
}

\end{table*}

\subsection{Results on Attack-Type Incremental Settings}

To further evaluate the robustness of our method, we evaluate it under attack-type incremental learning scenarios, referred to as Protocol-2. These settings simulate real-world conditions involving substantial domain gaps due to significant variations in attack types~\cite{wang2024multi}. Specifically, we construct subsets from the SiW and OULU-NPU datasets, each containing only one type of attack, namely Print or Replay, along with bona fide samples: SiW-Print (S-P), OULU-Replay (O-R), SiW-Replay (S-R), and OULU-Print (O-P). Two challenging sequences are designed to transition between attack types and datasets: SiW-Print → OULU-Replay (S-P→O-R) and OULU-Print → SiW-Replay (O-P→S-R).

As shown in Table~\ref{tab:cross_attack_results}, SVLP-IL demonstrates superior performance compared to both the PAD-specific incremental learning method MDIL-PAD~\cite{wang2024multi} and recent general incremental learning approaches such as SD-LoRA~\cite{wu2025sdlora} and KA-Prompt~\cite{xu2025kaprompt}. In the S-P$\to$O-R sequence, our method achieves an HTER of 0.00\% on the new target domain, O-R, notably surpassing even the Joint Training baseline of 0.28\%. Simultaneously, it maintains high stability on the initial S-P task with an HTER of 0.63\%, outperforming all comparison methods in mitigating forgetting.

In the O-P$\to$S-R sequence, SVLP-IL continues to exhibit the best trade-off between stability and plasticity. While SD-LoRA adapts well to the new task, it suffers from significant forgetting on the previous domain, yielding an HTER of 16.42\%. In contrast, SVLP-IL maintains a robust low HTER of 0.84\% on the original O-P task while effectively adapting to the new domain. Consequently, SVLP-IL yields the lowest $\Delta m\%$ values across both scenarios, specifically 0.18\% and 0.39\%, respectively, confirming the efficacy of its MAP and SEWC components in handling challenging incremental tasks with non-overlapping attack types.

\begin{table}[htbp]
\centering
\vspace{1mm}
% \caption{Performance comparison (HTER in \% and $\Delta m\%$) on cross-attack-type incremental learning settings (S-P~$\rightarrow$~O-R and O-P~$\rightarrow$~S-R). Lower values are better for both metrics.
% }
\caption{Cross-attack-type incremental learning performance in terms of HTER and $\Delta m\%$ on the S-P~$\rightarrow$~O-R and O-P~$\rightarrow$~S-R settings. Lower values are better for both metrics.
}
\label{tab:cross_attack_results}
\small
\resizebox{\linewidth}{!}{
\begin{tabular}{lccccccc}
\toprule
\multirow{2}{*}{\textbf{Methods}} & \multicolumn{3}{c}{\textbf{Step 2 (S-P→O-R)}} & \multicolumn{3}{c}{\textbf{Step 2 (O-P→S-R)}} \\
\cmidrule(r){2-4} \cmidrule(r){5-7}
 & S-P↓ & O-R↓ & $\Delta m\%$↓ & O-P↓ & S-R↓ & $\Delta m\%$↓ \\
\midrule
JT & 0.00 & 0.28 & - & 0.07 & 0.17 & - \\
FT & 3.67 & 1.55 & 2.47  & 14.18 & 0.13 & 7.04 \\
\midrule
MDIL-PAD~\cite{wang2024multi} & 0.65 & 0.63 & 0.50 & 6.65 & 0.28 & 3.35 \\
SD-LoRA~\cite{wu2025sdlora} & 3.50 & 0.91 & 2.07 & 16.42 & \textbf{0.00} & 8.10 \\
KA-Prompt~\cite{xu2025kaprompt} & 2.88 & 0.68 & 1.60 & 1.40 & 0.30 & 0.73 \\
\textbf{Ours} & \textbf{0.63} & \textbf{0.00} & \textbf{0.18} & \textbf{0.84} & 0.17 & \textbf{0.39} \\
\bottomrule
\end{tabular}
}

\end{table}

\subsection{Leave-One-Out Cross-Domain Generalization Analysis}

To assess SVLP-IL's generalization capability to entirely unseen domains, we conduct a leave-one-out experiment using MSU (M), CASIA (C), Idiap (I), and OULU (O) datasets, referred to as Protocol-3. Unlike conventional Domain Generalization (DG) methods that assume concurrent access to all source domains during training, our setting requires the model to learn from source domains incrementally and without replay, making the task considerably more difficult.

Performance metrics, namely are presented in Table~\ref{tab:exp3}. The results indicate that SVLP-IL performs favorably against both PAD-specific and general Incremental Learning (IL) baselines across the four leave-one-out folds. Specifically, SVLP-IL achieves the lowest HTERs among all IL methods when generalizing to domain C, I and O, achieving HTERs of 6.21\%, 7.83\%, and 8.77\%, respectively. Compared to the recent general IL approaches, our method demonstrates significantly better stability; for instance, on domain O, SVLP-IL outperforms SD-LoRA and KA-Prompt, which recorded 13.29\% and 10.80\% respectively, by a large margin. While MDIL-PAD achieves a slightly lower HTER on domain M, SVLP-IL outperforms it significantly on all other three domains, exemplified by the performance gap of 6.21\% versus 13.22\% on domain C, showcasing a more balanced and robust generalization ability.

Remarkably, despite the stricter rehearsal-free incremental constraints, SVLP-IL attains generalization performance competitive with, and in some cases superior to, several dedicated DG methods. For example, our HTER of 6.21\% on domain C is superior to SSDG-R~\cite{jia2020single} at 10.44\% and SSAN-R~\cite{wang2022domain} at 10.00\%, which have access to all source domains simultaneously. This indicates that the transferability of the representations learned through MAP and SEWC is robust enough to serve as a practical solution for real-world PAD systems requiring adaptation to unseen domains.

\begin{table*}[htbp]
\centering
\vspace{1mm}
% \caption{Leave-one-out cross-domain generalization performance (HTER in \% and AUC in \%). Models were incrementally trained on three source domains and evaluated on the fourth held-out target domain from the set \{MSU, CASIA, Idiap, OULU\}. Lower HTER and higher AUC indicate better generalization. Results for DA, DG, and IL baselines are included for comparison.}
\caption{Leave-one-out cross-domain generalization performance in terms of HTER and AUC. Models were incrementally trained on three source domains and evaluated on the fourth held-out target domain from the set \{MSU-MFSD, CASIA-MFSD, Idiap Replay-Attack, OULU-NPU\}. Lower HTER and higher AUC indicate better generalization.}
\label{tab:exp3}
\small
\resizebox{\textwidth}{!}{ % <-- 缩放表格
\begin{tabular}{llcccccccc}
\toprule
\textbf{} & \textbf{Methods} & \multicolumn{2}{c}{\textbf{O\&C\&I to M}} & \multicolumn{2}{c}{\textbf{O\&M\&I to C}} & \multicolumn{2}{c}{\textbf{O\&C\&M to I}} & \multicolumn{2}{c}{\textbf{I\&C\&M to O}} \\
\cmidrule(r){3-4} \cmidrule(r){5-6} \cmidrule(r){7-8} \cmidrule(r){9-10}
 & & HTER(\%)↓ & AUC(\%)↑ & HTER(\%)↓ & AUC(\%)↑ & HTER(\%)↓ & AUC(\%)↑ & HTER(\%)↓ & AUC(\%)↑ \\
\midrule
\multirow{3}{*}{DA} 
 & SDA~\cite{wang2021self} & 15.40 & 91.80 & 24.50 & 84.40 & 15.60 & 90.10 & 23.10 & 84.30 \\
 & VLAD~\cite{wang2021vlad} & 11.43 & 96.44 & 20.79 & 86.32 & 12.29 & 92.95 & 21.20 & 86.93 \\
 & GDA~\cite{zhou2022generative} & \textbf{9.20} & \textbf{98.00} & \textbf{12.20} & \textbf{93.00} & \textbf{10.00} & \textbf{96.00} & \textbf{14.40} & \textbf{92.60} \\
\midrule
\multirow{8}{*}{DG} 
 & MADDG~\cite{shao2019multi} & 17.69 & 88.06 & 24.50 & 84.51 & 22.19 & 84.99 & 27.98 & 80.02 \\
 & D2AM~\cite{chen2021generalizable} & 12.70 & 95.66 & 20.98 & 85.58 & 15.43 & 91.22 & 15.27 & 90.87 \\
 & FGHV~\cite{liu2022feature} & 9.17 & 96.92 & 12.47 & 93.47 & 16.29 & 88.79 & 13.58 & 93.55 \\
 & AMEL~\cite{zhou2022adaptive} & 10.23 & 96.62 & 11.88 & 94.39 & 18.60 & 88.79 & \textbf{11.31} & \textbf{93.96} \\
 & SSDG-R~\cite{jia2020single} & 7.38 & 97.17 & 10.44 & 95.94 & 11.71 & 96.63 & 15.61 & 91.54 \\
 & SSAN-R~\cite{wang2022domain} & \textbf{6.67} & \textbf{98.75} & \textbf{10.00} & \textbf{96.67} & \textbf{8.88} & \textbf{96.79} & 13.72 & 93.63 \\
\midrule
\multirow{4}{*}{IL} 
 & LwF~\cite{li2017learning}  & 17.14 & 98.18 & 33.94 & 69.56 & 20.29 & 91.24 & 19.03 & 88.93 \\
 & L2P~\cite{L2P_ref} & 13.57 & 93.23 & 22.28 & 83.45 & 11.82 & 95.25 & 31.74 & 76.08 \\
 & MDIL-PAD~\cite{wang2024multi} & \textbf{5.71} & \textbf{98.19} & 13.22 & 91.94 & 11.25 & 95.44 & 12.47 & 94.22 \\
 & SD-LoRA~\cite{wu2025sdlora} & 12.15 & 93.89 & 8.35 & 96.10 & 10.40 & 96.10 & 13.29 & 94.39 \\
 & KA-Prompt~\cite{xu2025kaprompt} & 14.97 & 90.01 &  6.22 &  98.05 & 13.76 & 94.75 & 10.80 & 95.44 \\
& \textbf{Ours} & 6.51 & 96.60 & \textbf{6.21} & \textbf{98.71} & \textbf{7.83} & \textbf{97.96} & \textbf{8.77} & \textbf{95.47} \\
\bottomrule
\end{tabular}
}

\end{table*}

\subsection{Performance on Long-Sequence Attack Increments and Generalization}

To simulate a more realistic lifelong learning scenario, we evaluate SVLP‑IL across a long sequence of eight diverse PAD datasets, referred to as Protocol‑4. The sequence is ordered by increasing attack complexity: MSU‑MFSD, CASIA‑FASD, Idiap REPLAY‑ATTACK, OULU‑NPU, SiW, ROSE‑YOUTU, HKBU‑MARs-V1+, and WFFD. We use AUC to track the per‑step trajectory across all learned domains, providing a threshold‑independent view of knowledge retention. Additionally, final generalization is assessed on the unseen CelebA‑Spoof dataset.

\begin{figure*}[htbp]
    \centering
    \includegraphics[width=\textwidth]{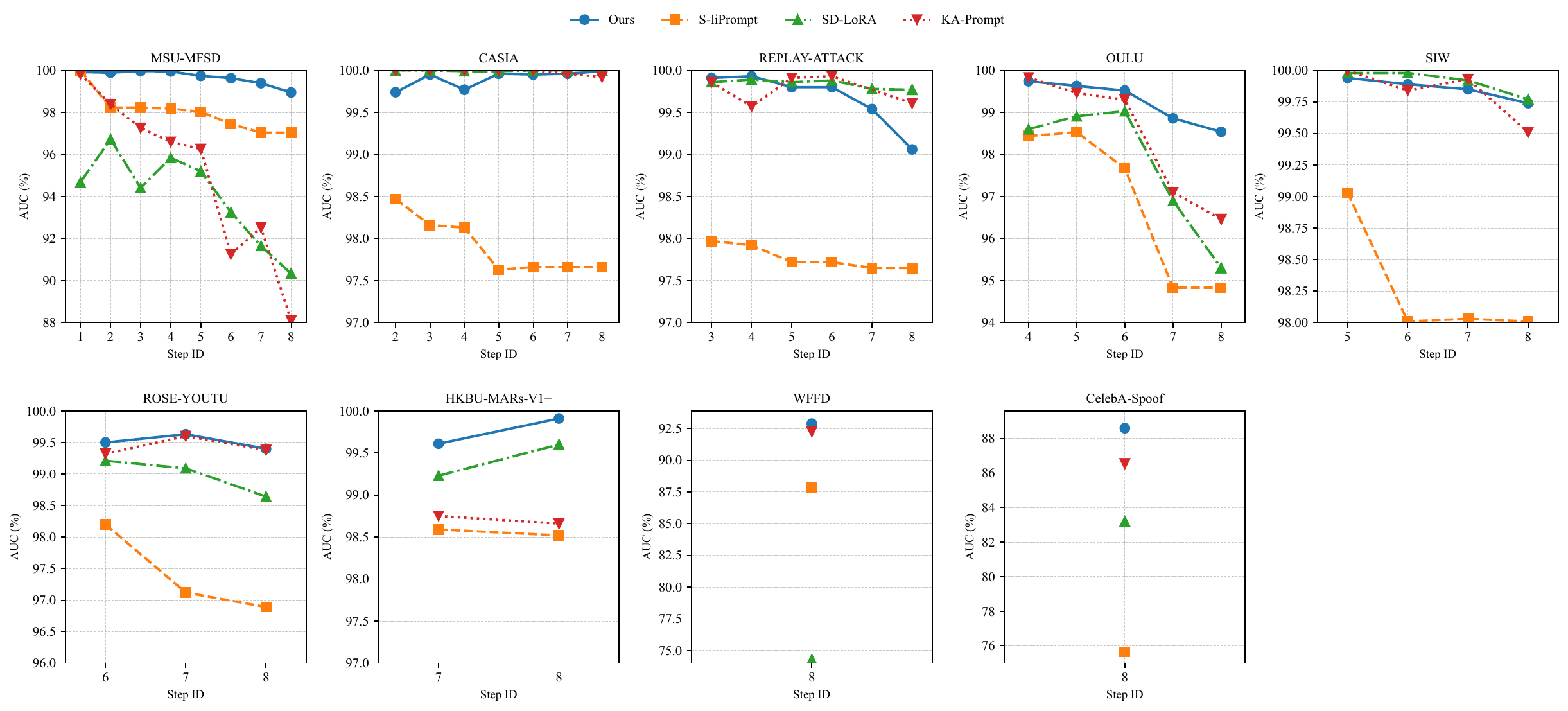}
    % \caption{Evaluation of incremental learning performance (AUC in \%) on Protocol-4, simulating progressively complex attack types. Subplots (MSU to WFFD) show performance on each source domain as new domains are sequentially learned (Step ID 1-8). The far-right subplot ("CelebA-Spoof") presents the final generalization AUC on an unseen diverse attack dataset after completing the 8-stage incremental training.}
    \caption{Long-sequence incremental learning performance in terms of AUC on Protocol-4, which simulates progressively complex attack types. Subplots (MSU-MFSD to WFFD) illustrate the performance trajectory on each source domain as new domains are sequentially learned. The last subplot (CelebA-Spoof) presents the final generalization performance on an unseen diverse attack dataset after completing the 8-stage incremental training.}
    \label{fig:exp4}
\end{figure*}

We compare our method against three representative approaches: S-liPrompts~\cite{wang2022_sprompts}, KA-Prompt~\cite{xu2025kaprompt}, and SD-LoRA~\cite{wu2025sdlora}. As shown in Fig.~\ref{fig:exp4}, each baseline reveals specific limitations when handling the complex evolution of spoofing attacks:

S-liPrompts relies on learning independent prompts for each domain while keeping the backbone frozen. This isolation prevents the transfer of shared spoofing cues across domains, unlike our MAP which captures universal knowledge. Furthermore, the frozen feature extractor fails to capture the fine-grained visual artifacts required for novel attack types. In contrast, our SVLP-IL allows controlled backbone updates via SEWC. This enables the model to learn new discriminative features for difficult attacks while protecting critical historical weights, resulting in significantly higher performance on early tasks like MSU compared to the rapidly decaying S-liPrompts.

KA-Prompt introduces component-wise alignment constraints to foster knowledge reuse. However, this strategy becomes a bottleneck when facing widely diverging attack types. Forcing alignment between structurally different attacks such as 2D prints and 3D wax figures leads to negative transfer. Consequently, KA-Prompt exhibits a clear downward trend in the later steps where attack complexity increases.

SD-LoRA displays marked instability characterized by a sawtooth volatility and suffers from plasticity exhaustion on the final WFFD task. Its low-rank adaptation mechanism restricts the parameter space available for new tasks. When facing highly diverse spoofing patterns, this constrained subspace becomes insufficient to represent new features without interfering with old ones, resulting in significant performance fluctuations and the failure to adapt to the final task.

In sharp contrast, SVLP-IL effectively overcomes these hurdles. By synergizing MAP for domain specialization and SEWC for stable backbone adaptation, our method maintains a flat and stable performance curve on the initial domain throughout the sequence. Furthermore, SVLP-IL retains high plasticity for late-stage adaptation and reaches an AUC of 92.89\% on WFFD, which is far superior to SD-LoRA. This balance culminates in superior generalization on the unseen CelebA-Spoof dataset, where SVLP-IL achieves the highest AUC of 88.59\%, outperforming KA-Prompt, SD-LoRA, and S-liPrompts.

\subsection{Robustness of the Domain Routing Mechanism}

We analyze the accuracy of our k-means domain router to understand its impact on model performance. We evaluate its classification accuracy at each incremental step of Protocol-1 for different numbers of prototypes, $k \in \{1, 3, 5, 7, 9\}$.

The results, shown in Table~\ref{tab:routing_accuracy_k}, reveal an important strength of our framework. While the routing accuracy is generally high, it is not perfect, with some misclassifications occurring as more domains are added. Despite these routing errors, our model maintains excellent end-to-end PAD performance, demonstrating a high degree of robustness.

This resilience is a direct benefit of the MAP design. The presence of shared domain-agnostic and mixed prompts ensures that the model receives reliable, universal spoofing cues even if the wrong domain-specific prompt is selected. This architecture makes the model tolerant to occasional routing imperfections.

\begin{table*}[ht]
\centering

\vspace{1mm}
% \caption{Routing accuracy (\%) of the k-means domain selection mechanism at each step of the Protocol-1 (M$\rightarrow$C$\rightarrow$I$\rightarrow$O) incremental learning sequence, evaluated for different numbers of prototypes ($k$).}
\caption{K-means domain selection performance in terms of routing accuracy on the Protocol-1 (M$\rightarrow$C$\rightarrow$I$\rightarrow$O) incremental learning sequence, evaluated across different numbers of prototypes ($k$).}

\label{tab:routing_accuracy_k}

\begin{tabular}{c|cc|ccc|cccc}
\toprule
 & \multicolumn{2}{c|}{\textbf{Step 2 (M$\rightarrow$C)}} & \multicolumn{3}{c|}{\textbf{Step 3 (M$\rightarrow$C$\rightarrow$I)}} & \multicolumn{4}{c}{\textbf{Step 4 (M$\rightarrow$C$\rightarrow$I$\rightarrow$O)}} \\
\cmidrule(lr){2-3} \cmidrule(lr){4-6} \cmidrule(lr){7-10}
\textbf{k} & \textbf{M}↑ & \textbf{C}↑ & \textbf{M}↑ & \textbf{C}↑ & \textbf{I}↑ & \textbf{M}↑ & \textbf{C}↑ & \textbf{I}↑ & \textbf{O}↑ \\
\midrule
1 & 95.27 & 94.84 & 79.73 & 93.41 & \textbf{87.90} & 72.30 & 94.27 & \textbf{90.01} & \textbf{89.27} \\
3 & 90.54 & 94.27 & 75.68 & 94.84 & 83.44 & 68.92 & 94.27 & 84.08 & 87.71 \\
5 & 97.30 & \textbf{99.43} & \textbf{86.49} & \textbf{97.99} & 83.86 & \textbf{78.38} & \textbf{96.56} & 83.01 & 89.05 \\
7 & 96.62 & 95.70 & 83.11 & 94.56 & 84.29 & 72.30 & 95.42 & 83.44 & 86.98 \\
9 & \textbf{98.65} & 94.56 & 83.78 & 95.42 & 86.20 & 72.97 & 95.42 & 86.20 & 87.93 \\
\bottomrule
\end{tabular}

\end{table*}

\subsection{Ablation Studies}
\subsubsection{\textbf{Ablation Study on MAP}}

\begin{figure}[t]
  \centering
  \includegraphics[width=0.9\linewidth]{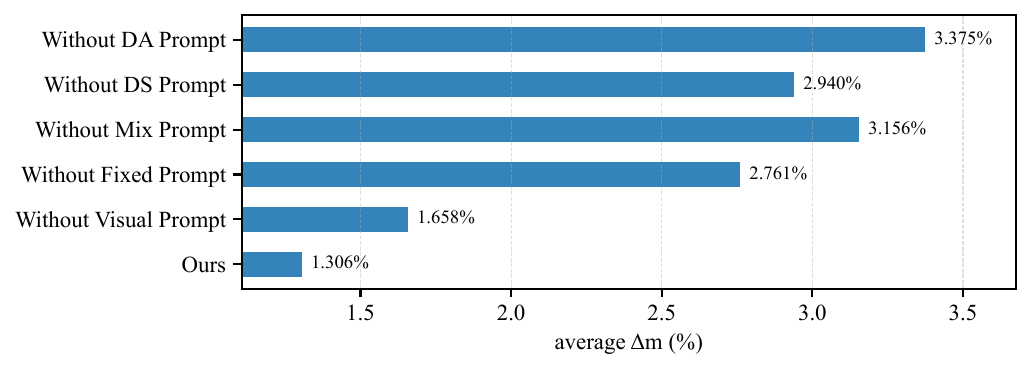} 
  \caption{Ablation study of Multi-Aspect Prompting (MAP) components on the long-sequence Protocol-4. We report the average $\Delta m\%$ over the 8 incremental steps. Lower values indicate better retention of past knowledge.}
  \label{fig:ablation_map}
\end{figure}

To rigorously evaluate the contribution of each component within our MAP framework, we conduct an ablation study using the challenging long-sequence Protocol-4. Instead of evaluating a single step, we report the average $\Delta m\%$ across all incremental steps. This metric provides a holistic view of the model's ability to retain knowledge over a prolonged lifecycle. The results are illustrated in Fig.~\ref{fig:ablation_map}.

The most critical degradation is observed when removing the Domain-Agnostic (DA) Prompt, with the average $\Delta m\%$ spiking to the highest value of 3.375\%. This underscores that the DA prompt serves as the primary repository for universal anti-spoofing knowledge, such as general skin texture analysis. Without this shared foundation, the model is forced to rely entirely on domain-specific parameters for each new task. This prevents the consolidation of cross-domain patterns, leading to severe catastrophic forgetting as the model effectively re-learns from scratch for each domain.

Removing the Domain-Specific (DS) Prompt also leads to a significant increase in forgetting, rising to 2.940\%. This result is particularly insightful: without DS prompts to absorb domain-specific noise, such as unique camera artifacts or lighting, the shared DA prompt is forced to overfit to the current domain's idiosyncrasies. This ``pollution'' of the shared knowledge base compromises its validity on previous domains. Similarly, the absence of the Mixed (Mix) Prompt increases the average $\Delta m\%$ to 3.156\%, breaking the bridge between general and specific knowledge, hindering the model's ability to balance common traits with local variations. Notably, removing the Fixed Prompt degrades stability, resulting in an average $\Delta m\%$ of 2.761\%. This confirms that the fixed natural language is necessary to leverage CLIP's zero-shot prior as a semantic anchor, preventing the learnable prompts from suffering semantic drift during optimization.

Finally, removing the Visual Prompt results in an average $\Delta m\%$ increase from 1.306\% to 1.658\%. By handling low-level domain shifts, such as resolution or color tone, at the input stage, visual prompts relieve the burden on the textual prompts. Without them, the textual prompts must over-compensate to handle visual discrepancies, leading to a distorted decision boundary that generalizes poorly to past domains. 

In summary, SVLP-IL achieves the lowest average $\Delta m\%$ of 1.306\% only when all components work in synergy, validating our hierarchical design.

\subsubsection{\textbf{Ablation Study on SEWC}}

\begin{figure}[!t]
    \centering
    \includegraphics[width=\linewidth]{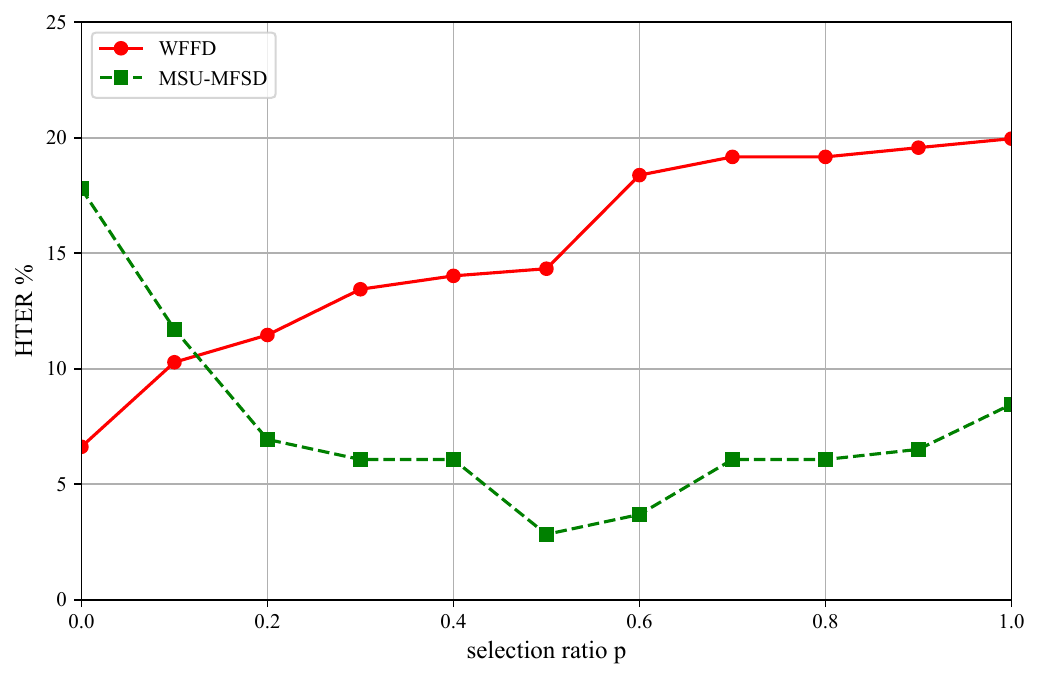}
    % \caption{Ablation study on the selection ratio $p$ in SEWC. HTER (\%) on WFFD (last domain) and MSU-MFSD (first domain) after sequential training on eight domains (Protocol-4).}
    \caption{Ablation study on the selection ratio $p$ in SEWC. Performance is evaluated in terms of HTER on WFFD (last domain) and MSU-MFSD (first domain) after sequential training on eight domains (Protocol-4).}
    \label{fig:asewc}
\end{figure}

To further analyze our Selective Elastic Weight Consolidation strategy, we evaluate how the parameter selection ratio $p$ affects the model's stability---defined as the retention of knowledge from early domains---and its plasticity---referring to its ability to adapt to new domains---in long-sequence incremental learning. Here, $p$ controls the proportion of parameters that are penalized, with a larger $p$ implying stronger consolidation by protecting a greater proportion of parameters. We sequentially train the model on eight domains under Protocol-4, while varying $p$ from $0$ to $1$ in increments of $0.1$, keeping all other configurations constant. The impact is measured by the HTER on both the first domain, MSU-MFSD, and the last domain, WFFD, after the incremental training is complete.

As shown in Fig.~\ref{fig:asewc}, the HTER on the new domain, WFFD, increases with $p$, indicating a loss of plasticity under stronger regularization. In contrast, the HTER on the old domain, MSU-MFSD, follows a U-shaped curve, initially decreasing but then rising as $p$ becomes large. This reveals that excessive regularization can also harm stability. The rationale is that a larger $p$ subjects a greater portion of the parameter space to constraints from all previously learned tasks. These conflicting constraints can ``pull'' the model parameters away from the optimal solution for MSU-MFSD, ultimately degrading performance on the very domain the strategy aims to protect.

This ablation underscores the importance of selective consolidation: SEWC outperforms both the naive, all-parameter EWC and the no-regularization baseline by focusing regularization on the most important parameters, thereby achieving a superior trade-off between stability and plasticity in incremental face anti-spoofing.

\section{Visualization and Analysis}

To visually elucidate how SVLP-IL mitigates catastrophic forgetting, i.e., particularly on domains where baseline methods struggle, we analyze Class Activation Maps (CAMs)~\cite{zhou2016learning_cam_ref}. We focus on a representative print attack from the MSU-MFSD dataset, as quantitative results show the Fine-tuning (FT) baseline exhibits severe forgetting on this domain in later stages of Protocol-1.

\begin{figure}[!t]
    \centering
    \includegraphics[width=\linewidth]{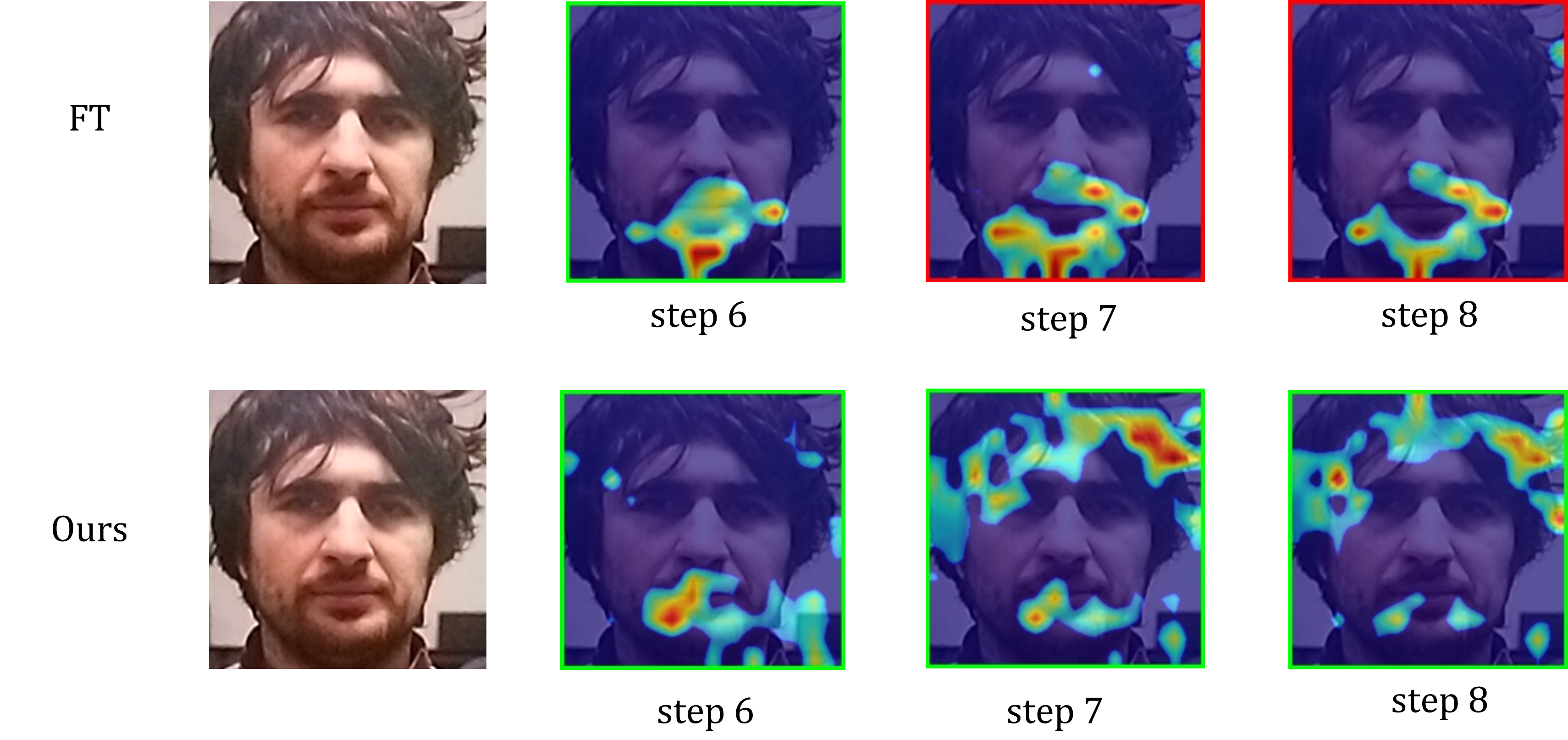}
    \caption{CAM visualizations comparing FT (top) and SVLP-IL (Ours, bottom) on a photo attack during incremental learning (step 6-8 of Protocol-4, where new attack types like 3D masks and wax figures were introduced in later steps). Green border: correct spoof detection. Red border: misclassified as real.}
    \label{fig:vis}
\end{figure}

Fig.~\ref{fig:vis} compares the CAM outputs of our method and FT across steps 6, 7 and 8 of Protocol-4. Steps 7 and 8 introduced 3D mask and wax figure attacks, respectively. Green borders indicate correct spoof classification while red denotes misclassification.

The FT baseline (top row) correctly classifies the print attack in step 6. However, after adapting to 3D masks in step 7, it misclassifies the same print attack as real, and this misclassification persists in step 8 after adapting to wax figure attacks.  The corresponding CAMs show that FT’s attention shifts away from print-specific artifacts toward general facial features, visually confirming that critical prior knowledge has been overwritten.

In contrast, SVLP-IL (bottom row) robustly classifies the print attack correctly throughout all steps. Its CAMs reveal stable, persistent attention to regions around the face periphery and its background interaction, i.e., areas often critical for detecting print artifacts, demonstrating clear resistance to catastrophic forgetting.

\section{Conclusion}
In this work, we introduce SVLP-IL, a novel rehearsal-free incremental learning framework that enables vision-language pre-trained models such as CLIP to continually adapt to face anti-spoofing (PAD) while effectively mitigating catastrophic forgetting. SVLP-IL achieves a robust balance between plasticity and stability without storing past data by synergistically combining a Multi-Aspect  Prompting (MAP) strategy for efficient domain adaptation with Selective Elastic Weight Consolidation (SEWC) to stabilize critical shared representations. Extensive experiments show that SVLP-IL significantly outperforms existing rehearsal-free methods in incremental PAD settings, demonstrating superior long-term stability and generalization. 

\nocite{*}
\bibliographystyle{IEEEtran}
\bibliography{references}
\ifCLASSOPTIONcaptionsoff
  \newpage
\fi

\end{document}